\documentclass{article}
\usepackage{ijcai20}

\usepackage{times}
\usepackage{soul}
\usepackage{url}
\usepackage[draft]{hyperref}
\usepackage[utf8]{inputenc}
\usepackage[small]{caption}
\usepackage{graphicx}
\usepackage{amsmath}
\usepackage{bm}
\usepackage{amsthm}
\usepackage{booktabs}
\usepackage{algorithm}
\usepackage{algorithmic}
\usepackage{graphicx}
\usepackage{epsfig}
\usepackage{color}
\usepackage{colortbl}
\usepackage{epstopdf}
\usepackage{caption}
\usepackage{multirow}
\usepackage{mathrsfs}
\usepackage{tabularx}
\usepackage{mathtools}
\usepackage{amssymb}
\usepackage{array,hhline}
\usepackage{threeparttable}
\usepackage{graphicx}
\usepackage{subcaption}
\usepackage{mwe}
\title{GoGNN: Graph of Graphs Neural Network for Predicting \\
Structured Entity Interactions}

\author{
Hanchen Wang$^1$\and
Defu Lian$^2$\footnote{Corresponding author}\and
Ying Zhang$^1$\and
Lu Qin$^1$ \and
Xuemin Lin$^3$
\affiliations
$^1$University of Technology Sydney\\
$^2$University of Science and Technology of China\\
$^3$University of New South Wales\\
\emails
hanchenw.au@gmail.com,
liandefu@ustc.edu.cn,\\
\{ying.zhang, lu.qin\}@uts.edu.au,
lxue@cse.unsw.edu.au
}

\begin{document}

\maketitle

\begin{abstract}
Entity interaction prediction is essential in many important applications such as chemistry, biology, material science, 
and medical science. The problem becomes quite challenging when each entity is represented by a complex structure,
namely structured entity, because two types of graphs are involved: local graphs for structured entities and a global graph to capture the interactions between structured entities. 
We observe that existing works on structured entity interaction prediction cannot properly exploit the unique graph of graphs model. In this paper, we propose a Graph of Graphs Neural Network, namely GoGNN, which extracts the features in both structured entity graphs and the entity interaction graph in a hierarchical way.
We also propose the dual-attention mechanism that enables the model to preserve the neighbor importance in both levels of graphs. Extensive experiments on real-world datasets show that GoGNN outperforms the state-of-the-art methods 
on two representative structured entity interaction prediction tasks: chemical-chemical interaction prediction and drug-drug interaction prediction. Our code is available at Github\footnote{\url{https://github.com/Hanchen-Wang/GoGNN}}.

\end{abstract}

\section{Introduction}

Interactions between the structured entities like chemicals are the basis of many applications such as chemistry, biology, material science, medical science, and environmental science.
For example, the knowledge of chemical interactions is a helpful guide for the toxicity prediction, new material design and pollutant removal~\cite{xu2019mr}. In medical science, understanding the interaction between drugs is vital for drug discovery and side effect prediction which can save millions of lives every year~\cite{menche2015uncovering}.

One immediate way to investigate the interactions between two structured entities
is to conduct experiments for them in the laboratory or clinics.
However, due to the enormous number of structured entities, it is infeasible in terms of both time and resource
to examine all possible interactions.
Thanks to the advances in the computational approaches for the structured entity interaction prediction,
a variety of techniques have been proposed to predict the interactions among structured entities effectively and efficiently by utilizing the deep neural network or graph neural network (GNN) techniques
such as DeepCCI~\cite{kwon2017deepcci} for chemical-chemical interaction prediction and
DeepDDI~\cite{ryu2018deep} for drug-drug interaction prediction.

\begin{figure}[tb]
\vspace{-2mm}
\centering
\includegraphics[width=\columnwidth]{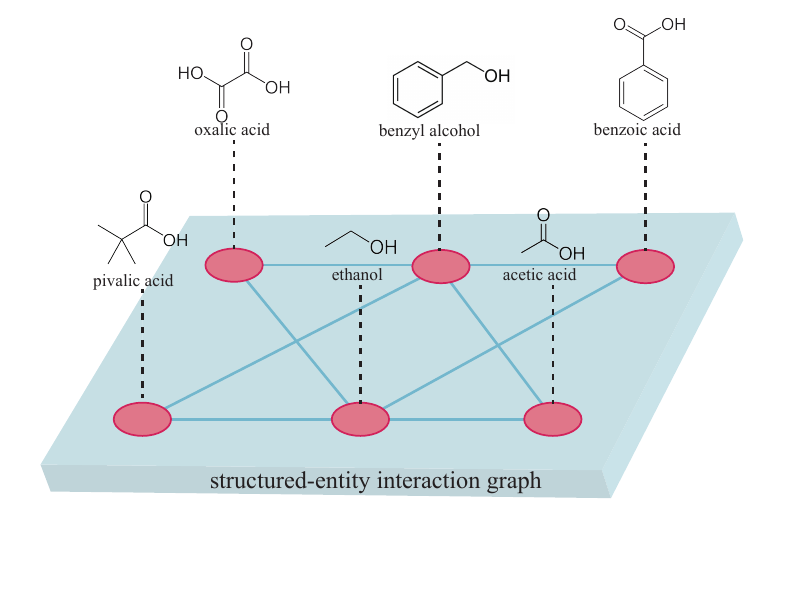}
\vspace{-12mm}
\caption{Interaction graph of molecule graphs.}
\vspace{-2mm}
\label{fig:intro}
\end{figure}

We observe that the interactions among structured entities can be naturally modeled by the graph-of-graphs (a.k.a network-of-networks) where each structured entity is a \textit{local} graph, and the interactions of the entities form a \textit{global} graph. In Figure~\ref{fig:intro}, we take the chemical-chemical interactions as an example.
Each chemical molecule is a structured entity and can be represented by a \textit{local} graph (i.e., molecule graph)
where nodes represent atoms and the bonds among the atoms are the edges.
On the other hand, the interactions (edges) among the structured entities (nodes) form a \textit{global} graph.
However, the existing studies for structured entity interaction prediction do not make full use of the graph-of-graphs model and only consider partial information.
For instance, MR-GNN \cite{xu2019mr} only considers the local structure information of entities
and their pairwise similarity; Decagon \cite{zitnik2018modeling} focuses on the interaction graph
and only treats the structured entity as a simple node.
Other works such as DeepCCI and DeepDDI even do not consider the graph structure information.

These limitations motivate us to develop a new approach to fully exploit the graph-of-graphs (GoG) model
to predict the structured entity interactions.
In particular, we propose a novel model called Graph of Graphs Neural Network({\bf GoGNN}).
Our model builds a graph neural network with attention-based pooling over local graphs and attention-based neighbor aggregation on the global graph such that GoGNN is able to capture broader information that enhances the performance on the prediction. Furthermore, the GNNs on both levels of graphs play synergistic effects on improving the representativeness of GoGNN.
The contributions of our model can be summarized as follows:

\begin{itemize}
\item To the best of our knowledge, this is the first work to systematically apply the graph neural network on graph-of-graphs model, namely Graph of Graphs Neural Network ({\bf GoGNN}),
to the problem of structured entity interaction prediction.

\item The proposed GoGNN mines the features from both local entity graphs and global interaction graph hierarchically and synergistically. We design dual attention architecture to capture the significance of the substructures in the local graphs while preserving the importance of the interactions within the global graph.

\item The extensive experiments conducted on the real-life benchmark datasets show that GoGNN outperforms the state-of-the-art structured entity interaction prediction methods in two representative applications: chemical-chemical interaction prediction and drug-drug interaction prediction.

\end{itemize}

\section{Related Work}

In this section, we introduce the closely related works.

\subsection{Structured Entities Interaction Prediction}
In many real-life applications such as chemistry, biology, material science, and medical science, we need to understand the interactions between the structured entities. 
In recent years, a variety of techniques have been proposed for structured entity interaction prediction in some specific applications. 
In this paper, we focus on two representative applications: chemical-chemical interaction prediction and drug-drug interaction prediction.

Many computational methods have been proposed for these two applications.
DeepCCI and DeepDDI~\cite{kwon2017deepcci,ryu2018deep} utilize the conventional convolutional neural network and PCA on the chemical data.
Some models are graph neural network-based.
For example, Decagon~\cite{zitnik2018modeling} performs the GCN on drug-protein interaction graph; MR-GNN~\cite{xu2019mr} proposes a model with dual graph-state LSTMs that extracts local features of molecule graphs, and MLRDA~\cite{chu2019mlrda} utilizes graph autoencoder with a novel loss function to predict the drug-drug interactions.


\subsection{Graph Neural Networks}
\vspace{1mm}
\noindent {\bf Node-level applications.}
Most GNNs are designed for node-level applications such as node classification and link prediction~\cite{DBLP:conf/iclr/KipfW17,velivckovic2017graph,zhang2018link,hamilton2017inductive,liu2019geniepath,10.1145/3366423.3380151,10.1145/3366423.3380187}.
They rely on the node embedding techniques like skip-gram, autoencoder and neighbor aggregation methods like GCN, GraphSAGE, etc. These methods focus on the node relations within the graph and use the low-dimension representations to preserve the structural and attribute information.

\vspace{1mm}
\noindent {\bf Graph-level applications.}
Recently, some research works on GNNs are proposed for graph-level applications such as graph classification\cite{zhang2018end,lee2018graph} and graph matching\cite{li2019graph}.
These works learn the graph representations for each graph individually or pair-wisely without considering the interactions between the graphs.

\subsection{Graph of Graphs}
In most real-world systems, an individual network is one component within a much larger complex multi-level network.
Applying the graph theory paradigm to these networks has led to the development of the concept of ``Graph of Graphs'' (also known as ``Network of Networks'').
\cite{d2014networks} introduces the theoretical research development~\cite{dong2013robustness}, applications~\cite{DBLP:conf/kdd/NiTFZ14} and phenomenological model~\cite{rome2014federated} on the network of networks.
These works enable us to understand and model the inter-dependent critical infrastructures.
SEAL\cite{DBLP:conf/www/LiRCMHH19} proposed graph neural network in a hierarchical graph perspective for graph classification task.
With significant differences between GoGNN and SEAL in tasks, loss functions and optimizers, GoGNN is the first work to develop graph neural network technique on graph of graphs for structured entity interaction prediction problem.


\section{Preliminaries}

\begin{figure*}[thb]
\centering
\includegraphics[width=2.1\columnwidth]{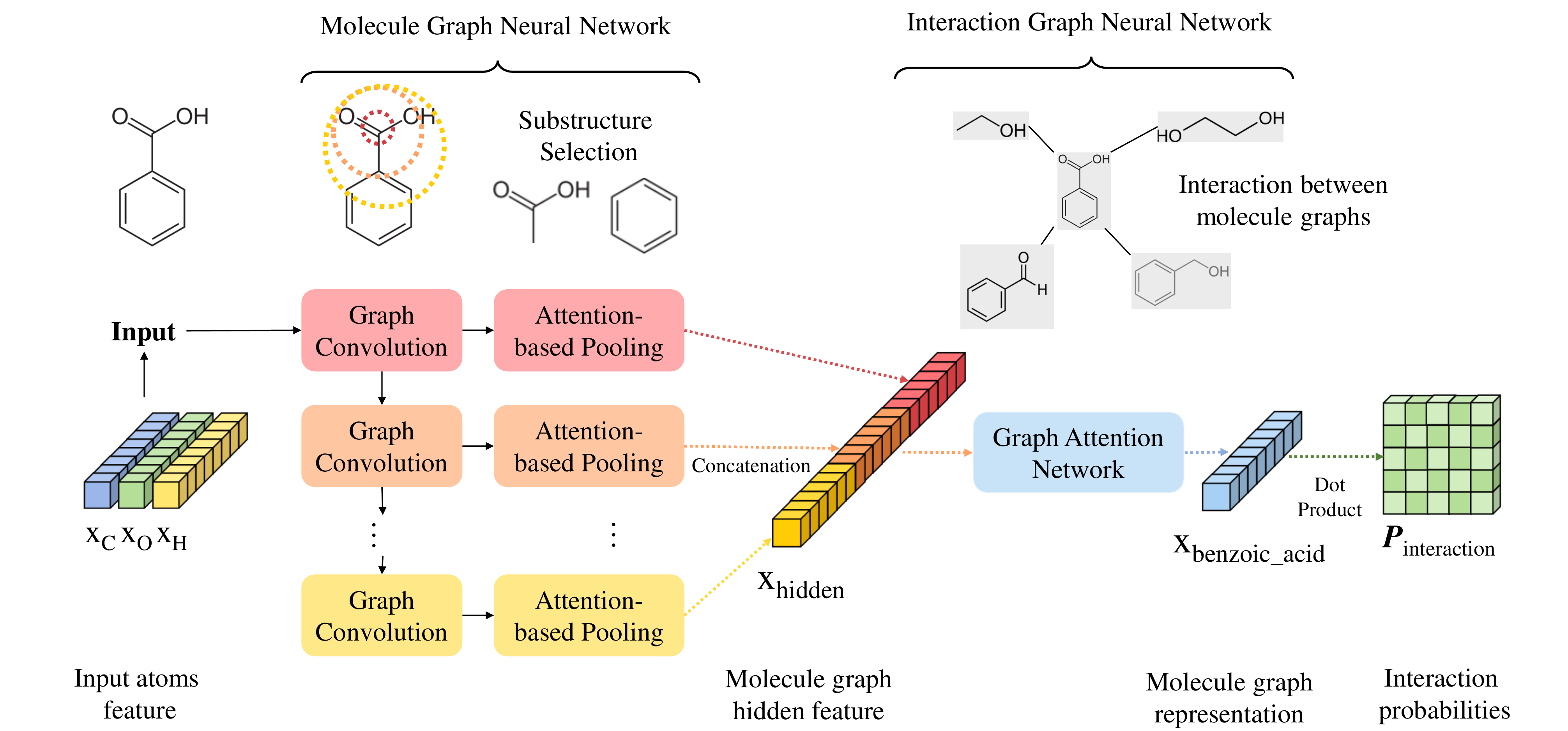}
\vspace{-5mm}
\caption{Framework of Graph of Graphs Neural Network.}
\vspace{-4mm}
\label{fig:frame}
\end{figure*}

\subsection{Problem Definition}
For ease of understanding of our techniques, in this paper, we focus on two representative applications of structured entities interaction prediction: chemical-chemical interaction (CCI) prediction and drug-drug interaction (DDI) prediction.
In the CCI graph, there is only one type of interaction, and our goal is to estimate the reaction probability score $p_{ij}$ of given chemical pair $(G_i, G_j)$.
As to DDI graph which has multiple types of interactions, we aim to estimate the occurrence probabilities $p^{r}_{ij}$ of side effect type $r$ with the given triplet $(G_i, r, G_j)$.

\subsection{Input Graph of Graphs}
Overall, the input interaction graph is regarded as graph-of-graphs as follows.

\vspace{1mm}
\noindent {\bf Molecule Graph.}
In both CCI and DDI prediction tasks, the local graphs are molecule graphs, each of which can be modeled as a heterogeneous graph with multiple types of nodes and edges.
In particular, the molecule graph $G_{M}$ consists of atoms $\{ a_i \}$ as nodes, and edges $\{ e_{ij} \}$ where $e_{ij}$ denotes the bond between atoms $a_i$ and $a_j$.
Each atom (i.e., node) $a$ is encoded as a vector ${\bm x}_a$.
For each bond (i.e., edge), we assign a weight to the corresponding edge depending on the type of the bond.
For example, the bond between the carbon atoms in the ethylene molecule is a double bond.
Therefore, the weight of the edge $e_{CC}$ between the carbon atoms is set to $2$.

\vspace{1mm}
\noindent {\bf Interaction Graph.}
The global interaction graph $G_{I}$ is formed by the molecule graphs and the interactions between them: $G_{I} = \{N, E_{I}\}$, where $N$ denotes the node set of $G_I$ which consists of molecule graphs $\{ G_{M} \}$, and $E_{I}$ denotes the interaction edges between the molecule graphs.
Note that, in CCI graph, there is only one type of interaction between two nodes.
In DDI graph, there are multiple types of side effects caused by the combination of two drugs.
An attribute vector ${\bm e}^r$ is assigned for each edge $e$ based on the side effect type $r$.

\section{Graph of Graphs Neural Network}
In this section, we introduce our Graph of Graphs Neural Network model.

\subsection{Framework of GoGNN}
The framework of GoGNN is illustrated in Figure~\ref{fig:frame}.
GoGNN contains molecule graph neural network which takes the atom features as input and interaction graph neural network which produces the graph representation for the prediction task. The two parts of GoGNN play a synergistic effect on improving the performance. The hidden features learned by molecule-level GNN provide the interaction-level GNN a representative initial input. The feature aggregation on the interaction-level GNN promotes the ability of molecule-level GNN to find key substructure through back-propagation.


\subsection{Molecule Graph Neural Network}
\label{sec:mol}
In organic chemistry, functional groups (i.e., substructures) in molecules are responsible for the characteristic chemical reactions between these molecules.
For example, the reaction between benzoic acid and ethanol in Figure~\ref{fig:intro} is
the esterification between two functional groups -COOH in benzoic acid  and -OH in ethanol.



The model could achieve better performance for prediction if the model can identify the functional groups in the molecules and represent the molecule with such functional groups.
Therefore, we designed our molecule graph neural network with the combination of multi-resolution architecture \cite{xu2019mr} which preserves the information of multi-hop substructures and attention-based graph pooling \cite{lee2019self,gao2019graph} which selects the substructures to represent the molecules.

As proved in previous work~\cite{xu2018powerful}, one single general graph convolution layer can only aggregate the feature of the node and its immediate neighbors. To obtain features of the multi-scale substructure of the molecule graph,  we apply multiple layers of graph convolution operations to the input graphs.
The graph convolution operation at $l^{th}$ layer is summarized as follows

\begin{equation}
\label{GCN}
\begin{aligned}
    {\bm M}_{(l+1)} &= GCN_{l}({\bm A}, {\bm M}_{l}) \\
    GCN_{l}({\bm A}, {\bm M}_{l}) &= \sigma(\tilde{\bm D}^{-\frac{1}{2}}\tilde{{\bm A}}\tilde{\bm D}^{-\frac{1}{2}}{\bm M}_{l}{\bm W}_{l})
\end{aligned}
\end{equation}

where ${\bm M}_{l} \in \mathbb{R}^{n \times d}$ is the hidden feature matrix for molecule graph at $l^{th}$ layer, $\tilde{\bm A} = {\bm A}+{\bm I}$ is the adjacency matrix with self-connection for molecule graph $G_{M}$, $\tilde{\bm D}$ is the diagonal degree matrix of $\tilde{\bm A}$ and $\sigma(\cdot)$ denotes the activation function.

Different from MR-GNN \cite{xu2019mr} which uses dual graph-state LSTMs on the input of subgraph representations, GoGNN applies graph pooling for learning the graph representation that preserves the substructure information,
in order to reduce the time and space complexity significantly.
As shown in Figure~\ref{fig:frame}, the self-attention graph pooling layer takes the output of each graph convolution layer as input to select the most representative substructures (functional groups) by learning the self-attention score ${\bm s}_l \in \mathbb{R}^{n \times 1}$ for molecule graph $G_{M}$ with $n$ atoms at $l^{th}$ layer

\begin{equation}
\label{score}
{\bm s}_l = \sigma(\tilde{\bm D}^{-\frac{1}{2}}\tilde{\bm A}\tilde{\bm D}^{-\frac{1}{2}}{\bm M}_{l}{\bm W}_{att}^{l})
\end{equation}

where ${\bm W}_{att}^{l} \in \mathbb{R}^{d \times 1}$ is the attention weight matrix for the pooling layer to obtain the self-attention score.
In order to select the most representative substructure, the graph pooling layer calculates the attention score for each atom in the graph and finds the top-$\lceil \gamma n \rceil$ atoms with the highest attention scores.
We set a hyperparameter pooling ratio $\gamma \in (0,1]$ to determine the number of nodes $\lceil \gamma n \rceil$ that are selected to represent the molecule graph

\begin{equation}
\begin{aligned}
idx = top({\bm s},& \lceil \gamma n \rceil),
{\bm s}_{mask} = {\bm s}_{idx}\\
{\bm M}_{sel} &= {\bm M} \odot {\bm s}_{mask}
\end{aligned}
\end{equation}

where $top$ is the function that returns the indices of atoms with top $\lceil \gamma n \rceil$ attention scores as in~\cite{DBLP:conf/icml/GaoJ19}; ${\bm s}_{mask} \in \{0, 1\}^{n \times 1}$ is the mask vector determined by the attention score;
$\odot$ denotes the column-wise product for masking;
${\bm M}_{sel}$ is the feature matrix of selected atoms in a molecule graph.
Afterward, the readout layer, which contains mean and sum pooling, is applied on the embedding of selected atoms ${\bm M}_{sel}$ to produce the molecule graph hidden feature.
After multiple graph convolutional and self-attention graph pooling layers, we got several graph hidden features.
Once obtained, we concatenate the outputs of the graph pooling layers as the hidden feature vector ${\bm x}_{G_{M}}$ for the molecule graph.
Because the hierarchical graph pooling architecture is applied, the graph representation can preserve the multi-hop substructure information effectively.
Hence, GoGNN can identify the function groups which play the key roles in molecule interactions and use these functional groups to represent the molecule graph.

\subsection{Interaction Graph Neural Network}
Most of existing CCI and DDI prediction models train the model with the input of pair of molecule graphs, but ignore the molecule interaction graph.
However, the information of interaction graph is crucial for the interaction prediction because it enables the model to capture high-order interaction relationship and enhance the model's ability to capture the representative molecular substructures synergistically.

We have the following observations that motivate us to perform graph neural network on the interaction graph:
Firstly, the type of interaction is dependent on the type of involved molecules.
As mentioned in Section~\ref{sec:mol}, esterification is the reaction between -OH in alcohols and -COOH in carboxyl acids.
The neighbor aggregation of GNN can gather the neighbor information that helps to summarize the types of chemicals that interact with the selected one.
Secondly, it is necessary to assign importance score to the neighbors for molecules in the interaction graph, since the chemical interactions have different significance and frequency.
For example, vitamin C has two main properties: reducibility and acidity.
Therefore, vitamin C cannot be prescribed with oxidizing drugs like vitamin K1 and alkaline drugs like omeprazole.
In an uncommon case, vitamin C reduces the therapeutic effect of inosine because of their complex physical and chemical reactions.
Therefore, we apply the graph attention network in order to preserve the frequencies of the chemical reactions and reduce the influence of biased observation of the interaction graph.
As for the DDI graph with edge attributes, an edge-aggregation graph neural network is applied.


\vspace{1mm}
\noindent {\bf Graph Attention Network.}
The attention-based graph neural network \cite{velivckovic2017graph} is applied on the interaction graph without edge attributes.
With the learned molecule hidden feature vector ${\bm x}_{G_{M}}$ and interaction graph $G_{I} = \{N, E_{I}\}$ as input, molecule graph representations are calculated by the neighbor aggregation on the interaction graph as follows

\begin{equation}
{\bm x}^{l+1}_{G_{i}} =  \mathbin{\Vert}^{K}_{\kappa =1}\sigma(\sum\limits_{j \in \eta_{G_{i}}}\alpha^{\kappa}_{ij}{\bm W}^{l}_{\kappa}{\bm x}^{l}_{G_{j}})
\end{equation}

where $K$ is the number of attention heads, $\sigma$ is a nonlinearity function, ${\bm W}^{l}_{\kappa}$ is the weight matrix at $\kappa^{th}$ attention head in $l^{th}$ layer and $\eta_{G_{i}}$ is the  set of neighbor molecule graphs of $G_{i}$ in the interaction graph $G_{I}$.
Notation $\alpha^{\kappa}_{ij}$ is the attention coefficient between $G_{i}$ and $G_{j}$ which is calculated by the following equation:

\begin{equation}
\alpha_{ij} = \frac{exp({\rm LeakeyRelu}({\bm a}[{\bm W}{\bm x}_{G_{i}}]\mathbin{\Vert}[{\bm W}{\bm x}_{G_{j}}]))}{\Sigma_{n \in \eta_{G_{i}}}exp({\rm LeakeyRelu}({\bm a}[{\bm W}{\bm x}_{G_{i}}]\mathbin{\Vert}[{\bm W}{\bm x}_{G_{n}}]))}
\end{equation}

where ${\bm a}$ is a learnable attention weight vector and $\mathbin{\Vert}$ is the concatenation operation.

\vspace{1mm}
\noindent {\bf Edge Aggregation Network.}
In DDI graph, each edge has an attribute vector ${\bm e}^{r}_{ij}$ which is determined by the side effect type $r$ of the drug combination $(G_i,  G_j)$.
To capture the edge attributes~\cite{schlichtkrull2018modeling}, we propose an edge aggregation network that aggregates the neighbor information together with edge attribute:

\begin{equation}
{\bm x}^{l+1}_{G_{i}} = \sigma({\bm W}^l{\bm x}^{l}_{G_i} + \sum\limits_{r}(\sum\limits_{{G_j}\in \eta^{r}_{G_i}} {\bm x}^{l}_{G_j}\cdot h_{{\bm W}_{\bm e}}({\bm e}^{r}_{ij})))
\end{equation}

where $h_{{\bm W}_{\bm e}}$ is the MLP layer with linear transformation matrix ${{\bm W}_{\bm e}}$ which transforms the edge attribute vector
${\bm e}^{r}_{ij} \in \mathbb{R}^{h \times 1}$ into a real number $\tau^{r}_{ij} \in  \mathbb{R}$.
In this way, GoGNN aggregates node's neighbor information together with edge attributes.
Different from Decagon \cite{zitnik2018modeling} which sets side-effect-specific parameters, GoGNN shares the parameters for all types of side effects in order to improve the robustness and generalization of the model.


\subsection{GoGNN Model Training}
We optimize the parameters with the task-specific loss functions.

\vspace{1mm}
\noindent {\bf Chemical Interaction Prediction.}
Since there is no edge attribute in the graph, we regard the chemical interaction prediction as a link prediction problem.
The dot product of two graph representations is used as the link probability of two graphs:
\begin{equation}
p_{ij} = \sigma({\bm x}^T_{G_i} \cdot {\bm x}_{G_j})
\end{equation}
where $\sigma$ is the activation function such as \textit{sigmoid} function that ensures $p_{ij} \in (0,1)$.
To encourage the model to assign higher probabilities to the observed edges than the random non-edges, we follow the previous study and estimate the model through negative sampling.
For each positive edge pair ($G_i$, $G_j$), a random negative edge ($G_i$, $G_m$) is sampled by choosing a molecule graph $G_m$ randomly.
We optimize the model using the following cross-entropy loss function
\begin{equation}
\mathcal{L}_{CCI} = \sum\limits_{(G_i, G_j) \in G_{CCI}}-log(p_{ij}) - \mathbb{E}_{m \sim P_j}log(1-p_{im})
\end{equation}

\vspace{1mm}
\noindent {\bf Drug Interaction Prediction.}
The drug-drug interaction prediction task is regarded as a multirelational link prediction problem.
Inspired by the loss design in \cite{zitnik2018modeling}, we train the parameters with the following cross-entropy loss function
\begin{equation}
p^{r}_{ij} = \sigma(({\bm W}_{r}{\bm x}_{G_i})^T \cdot ({\bm W}_{r}{\bm x}_{G_j}))
\end{equation}
\begin{equation}
\mathcal{L}^{r}_{ij} = -log(p^{r}_{ij}) - \mathbb{E}_{m \sim P^{r}_j}log(1-p^{r}_{im})
\end{equation}
\begin{equation}
\mathcal{L}_{DDI}  =\sum\limits_{(G_i, r, G_j) \in G_{DDI}} \mathcal{L}^{r}_{ij}
\end{equation}
where ${\bm W}_{r}$ is the side-effect-specific weight for linear transformation of ${\bm x}_{G_i}$ w.r.t. the side effect type $r$.
Given observed triplet $(G_i, r, G_j)$, the negative sample is chosen by replacing $G_j$ with randomly selected graph $G_m$ according to sampling distribution $P^{r}_j$~\cite{mikolov2013distributed}. 
\section{Experiment}
In this section, we introduce the extensive experiment results that demonstrate the effectiveness and robustness of GoGNN.

%
%

\subsection{Dataset}
\label{dataset}
To test the performance of our model on chemical-chemical interaction and drug-drug interaction prediction tasks, following datasets are chosen for the experiments:

\vspace{1mm}
\noindent {\bf CCI.}
The CCI dataset\footnote{\url{http://stitch.embl.de/download/chemical_chemical.links.detailed.v5.0.tsv.gz}} assigns a score from 0 to 999 to describe the interaction probability where a higher score indicates higher interaction probability.
According to threshold score, we get two datasets with chemical interaction probability score over 900 and 950:  CCI900 and CCI950. 
CCI900 has 14343 chemicals and 110078 chemical interaction edges, and CCI950 has 7606 chemicals and 34412 chemical interaction edges.

\vspace{1mm}
\noindent {\bf DDI.} For the drug-drug interaction prediction problem, DDI dataset\footnote{\url{https://www.pnas.org/content/suppl/2018/04/14/1803294115.DCSupplemental}} and the side effect dataset SE\footnote{\url{http://snap.stanford.edu/decagon}}~\cite{zitnik2018modeling} are used.
The DDI dataset is proposed by DeepDDI~\cite{ryu2018deep} which contains 86 types of side effects, 1704 drugs and 191400 drug interaction edges.
SE dataset is the integration of SIDER (Side Effect Resource), OFFSIDES and TWOSIDES database.
To familiarize the comparison,  we use the preprocessed data used by  Decagon~\cite{zitnik2018modeling}.
Therefore, the SE dataset contains 645 drugs, 964 types of side effects and 4651131 drug-drug interaction edges.
A vector representation ${\bm se}_r \in \mathcal{R}^{128}$ is assigned to each side effect type produced by pre-trained BERT model~\cite{devlin2018bert}.

The molecules are transformed from the SMILE strings~\cite{weininger1989smiles} into graphs by the open-source rdkit~\cite{landrum2013rdkit}.
An initial feature vector ${\bm x}_a \in \mathcal{R}^{32}$ is assigned for every atom.
The edges in molecule graphs are weighted by the type of the bonds.

\subsection{Baselines}
The proposed GoGNN is compared with the following state-of-the-art models:

\noindent{\bf DeepCCI}~\cite{kwon2017deepcci} is the CNN based model for predicting the interactions between the chemicals.

\noindent{\bf DeepDDI}~\cite{ryu2018deep} is the model designs a feature called structural similarity profile(SSP) combined with traditional MLP for DDI prediction.

\noindent{\bf Decagon}~\cite{zitnik2018modeling} is a GCN model on the drug and protein interaction graphs to predict the polypharmacy side effects caused by drug combinations. 

\noindent{\bf MR-GNN}~\cite{xu2019mr} is an end-to-end graph neural network with multi-resolution architecture that produces interaction between pairs of chemical graphs.

\noindent{\bf MLRDA}~\cite{chu2019mlrda} is the multitask, semi-supervised model for DDI prediction.

\noindent{\bf SEAL}~\cite{DBLP:conf/www/LiRCMHH19} is the neural network on hierarchical graphs for graph classification.

We used the public code of the baselines and keep the settings of models the same as mentioned in the original papers. We reimplemented SEAL for CCI and DDI prediction.

\vspace{1mm}
\noindent {\bf Ablation Study}

To investigate how the graph of graphs architecture and dual-attention mechanism improve the performance of the proposed model, we conduct the ablation study on the following variants of GoGNN:

\noindent {\bf GoGNN-M} is the variant which only learns the representations for the molecule-level graphs without the graph convolution on the interaction graph.
An MLP layer is applied with the input of molecule-level graph representations for the graph interaction prediction task.

\noindent {\bf GoGNN-I} only conducts graph convolution operation on the chemical interaction graphs. 
The initial molecule representations are the sum pooling of the atom representations within the molecule.

\noindent {\bf GoGNN-noPool} replaces the self-attention pooling on the molecule graph by the concatenation of conventional mean pooling and sum pooling.

\noindent {\bf GoGNN-noAttn} replaces the attention-based neural network on the interaction graph by a conventional GCN.

\subsection{CCI Prediction Results}
\label{sec:cci_pred}

\vspace{1mm}
\noindent {\bf Settings.}
Following the previous study, we divide the CCI datasets into training and testing set with ratio 9:1, and randomly choose 10\% data for validation.
The dimensions of molecule graph hidden feature, and the output molecule graph representation are set to 384, 256, respectively.
We set the learning rate to 0.01 and the pooling ratio to 0.5.
To evaluate the performance, we choose \textit{area under the ROC curve}(\textit{AUC}) and \textit{average precision score}(\textit{AP}) as metrics.

\begin{table}[tb]
\centering
\begin{tabular}{c|cccc}
\toprule[1pt]
& \multicolumn{2}{c}{\bf CCI900}&\multicolumn{2}{c}{\bf CCI950}\\
\midrule
 &AUC&AP&AUC&AP\\
\midrule
DeepCCI & 0.925&0.918&0.957&0.957\\
DeepDDI & 0.891 & 0.886&0.916&0.915\\
MR-GNN & 0.927 & 0.921 & 0.934 & 0.924\\
MLRDA & 0.922&0.907&0.959&0.948\\
SEAL & 0.894 & 0.886 &0.941 & 0.937\\
{\bf GoGNN}& {\bf 0.937} & {\bf 0.932} & {\bf 0.963} & {\bf 0.962}\\
\midrule
GoGNN-M &0.914 & 0.909 &0.938&0.931\\
GoGNN-I &0.921&0.898&0.929&0.912\\
GoGNN-noPool &0.931&0.930&0.958&0.954\\
GoGNN-noAttn &0.909&0.905&0.956&0.948\\
\bottomrule[1pt]
\end{tabular}
\vspace{-2mm}
\caption{Result of chemical-chemical interaction prediction task. }
\vspace{-1mm}
\label{tab:cci}
\end{table}

\vspace{1mm}
\noindent {\bf Results.}
As shown in Table~\ref{tab:cci}, GoGNN outperforms all the other state-of-the-art baseline methods on the CCI prediction task.
The improvement indicates that, compared with the methods that only train the parameters with pair-wise or individual chemical inputs, GoGNN can preserve more useful information on different scales by the feature extraction and aggregation through the graph of graphs.
The dual-attention mechanism also helps the model to learn higher quality graph representations by identifying and preserving the importance of molecular substructures and chemical interactions.

\subsection{DDI Prediction Results}
\vspace{1mm}
\noindent {\bf Settings.}
To familiarize the comparison, we divide the DDI dataset for training, testing, validation with ratio 6:2:2, and divide the SE dataset with ratio 8:1:1.
The dimensions of molecule graph hidden feature, and the output molecule graph representation are set to 384, 256, respectively.
We set the learning rate to 0.001, pooling ratio $\tau=0.5$.
We choose \textit{AUC} and \textit{average precision}(\textit{AP}) for evaluation.

\begin{table}[tb]
\centering
\begin{threeparttable}
\begin{tabular}{c|cccc}
\toprule[1pt]
& \multicolumn{2}{c}{\bf DDI}&\multicolumn{2}{c}{\bf SE}\\
\midrule
 &AUC&AP&AUC&AP\\
\midrule
DeepCCI & 0.862&0.856&0.819&0.806\\
DeepDDI & 0.915 & 0.912&0.827&0.809\\
MR-GNN & 0.932 & 0.922 & $0.769^*$ & $0.752^*$\\
MLRDA & 0.931&0.926&$0.847^*$&$0.825^*$\\
Decagon &-&-&0.872&0.832\\
SEAL &0.925&0.921&N/A&N/A\\
{\bf GoGNN}& {\bf 0.943} & {\bf 0.933} & {\bf 0.930} & {\bf 0.927}\\
\midrule
GoGNN-M &0.905 & 0.902 &0.862&0.817\\
GoGNN-I &0.922&0.917&0.860&0.834\\
GoGNN-noPool &0.900&0.891&0.912&0.909\\
GoGNN-noAttn &0.925&0.921&0.897&0.883\\
\bottomrule[1pt]
\end{tabular}

\begin{tablenotes}
\item[*] indicates that the result is the output of the baselines after two weeks' training.
\item[-] DDI dataset has no protein data which is required by Decagon
\end{tablenotes}
\end{threeparttable}
\vspace{-2mm}
\caption{Result of drug-drug interaction prediction task. }
\label{tab:se}
\end{table}

\vspace{1mm}
\noindent {\bf Results.}
The experiment results for DDI prediction are listed in Table~\ref{tab:se}.
The results show that compared with the baseline methods, GoGNN improves the performance with a significantly large margin.
GoGNN improves the AUC and AP by 1.18\% and 1.19\% respectively on DDI dataset, and 6.65\% and 11.42\% respectively on the SE dataset.
The improvement is attributed to the abundant information brought by the graph of graphs architecture and edge-filtered aggregation.

\subsection{Ablation Experiments}
The ablation experiment results on both tasks are shown in Table~\ref{tab:cci} and Table~\ref{tab:se}.
The results prove that the graph of graphs architecture, attention-based pooling, attention-based and edge-filtered aggregation are all effective for the side effect prediction task.
Among all the variants, GoGNN-M and GoGNN-I have the most significant performance gaps between GoGNN, which indicates that the view of graph of graphs contributes the most to helping the model to capture more structural information that improves the prediction accuracy.

\subsection{Parameter Sensitivity Analysis}
In this experiment, we test the impact of the hyper-parameters of GoGNN.

\vspace{1mm}
\noindent {\bf Settings.}
We conduct the parameter sensitivity experiment on the CCI950 dataset by changing the tested hyper-parameter while keeping other settings the same as mentioned in Section~\ref{sec:cci_pred}.
We test the following hyper-parameters: the dimensions of the output representation and hidden feature, learning rate and pooling ratio.

\begin{figure}
        \centering
        \begin{subfigure}[b]{0.475\columnwidth}
            \centering
            \includegraphics[width=\columnwidth]{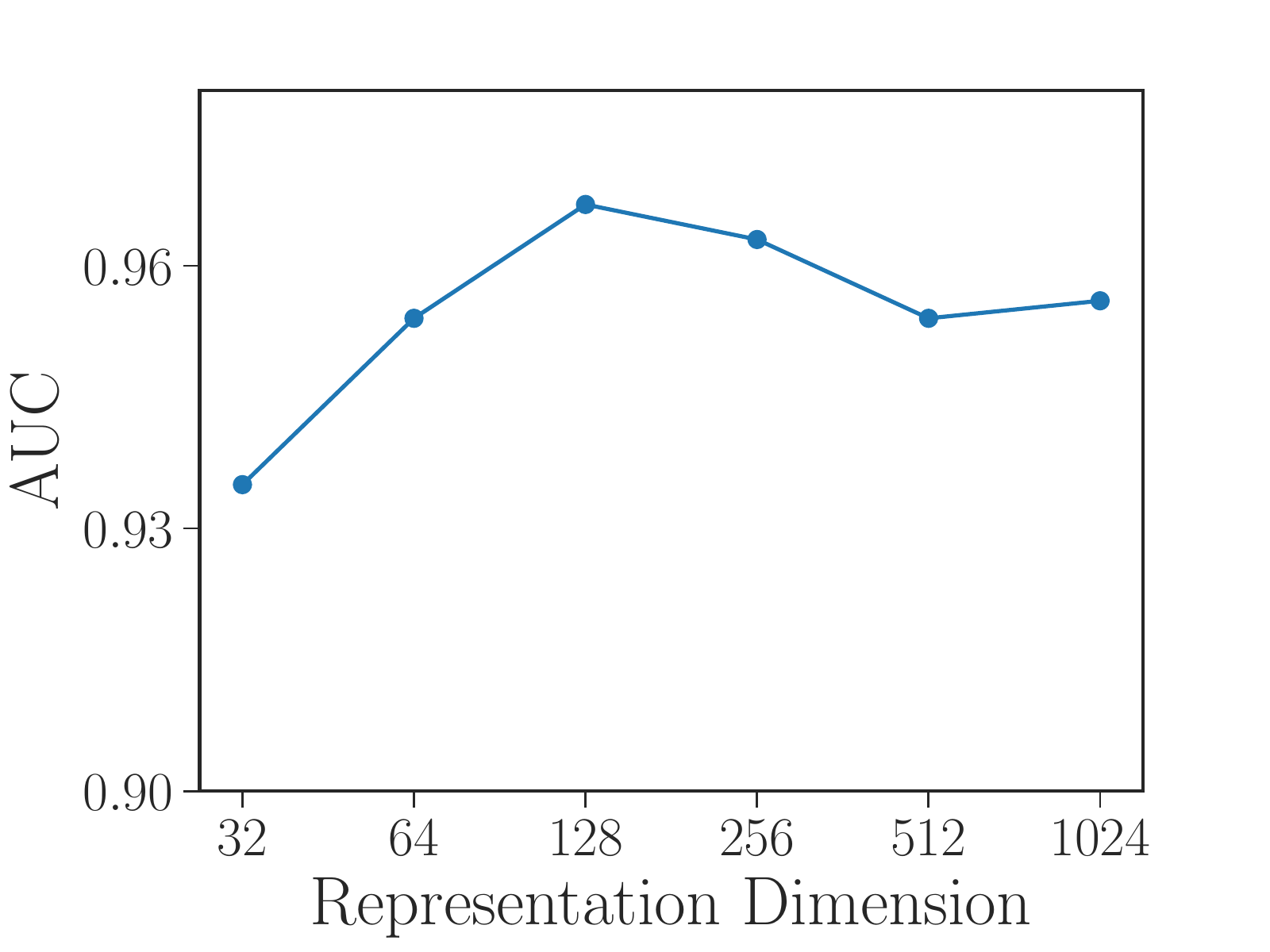}
            \caption[Network2]%
            {}    
            \label{fig:d_emb} 
\vspace{-4mm}
        \end{subfigure}
        \hfill
        \begin{subfigure}[b]{0.475\columnwidth}  
            \centering 
            \includegraphics[width=\columnwidth]{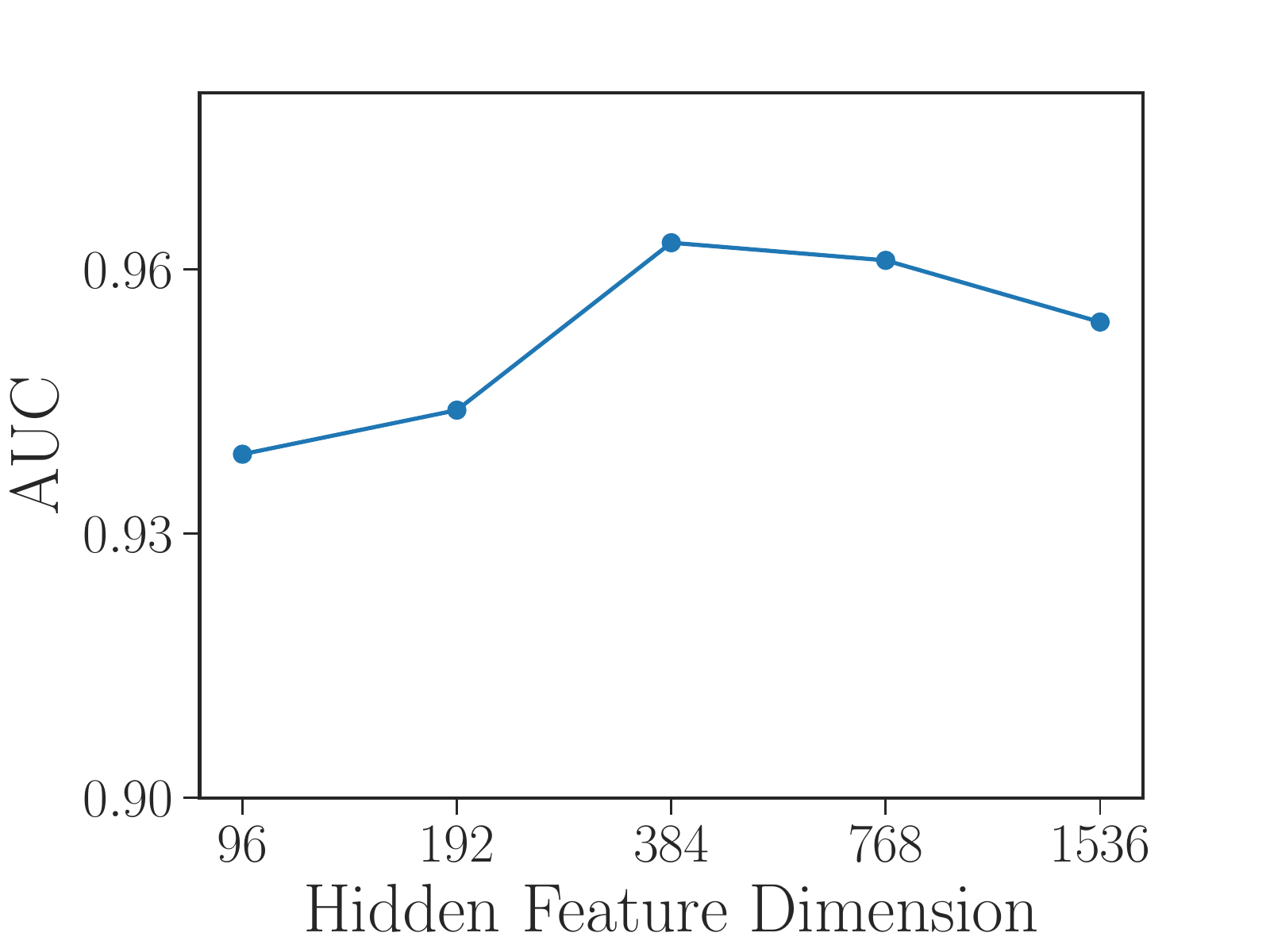}
            \caption[]
            {}    
\vspace{-4mm}
            \label{fig:d_hid}
        \end{subfigure}
        \vskip\baselineskip
        \begin{subfigure}[b]{0.475\columnwidth}   
            \centering 
            \includegraphics[width=\columnwidth]{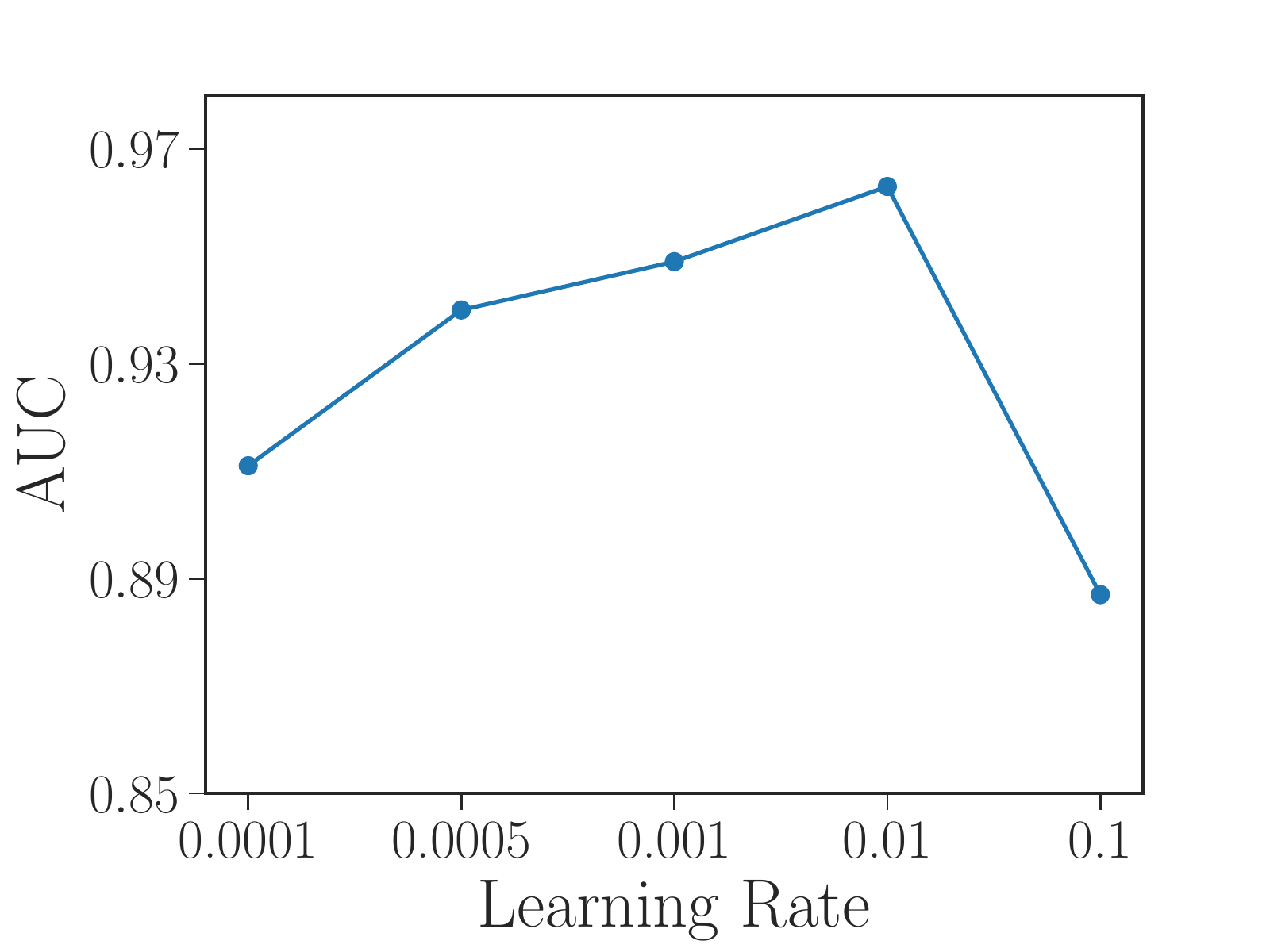}
            \caption[]%
            {}   
\vspace{-2mm}
            \label{fig:lr}
        \end{subfigure}
        \hfill
        \begin{subfigure}[b]{0.475\columnwidth}   
            \centering 
            \includegraphics[width=\columnwidth]{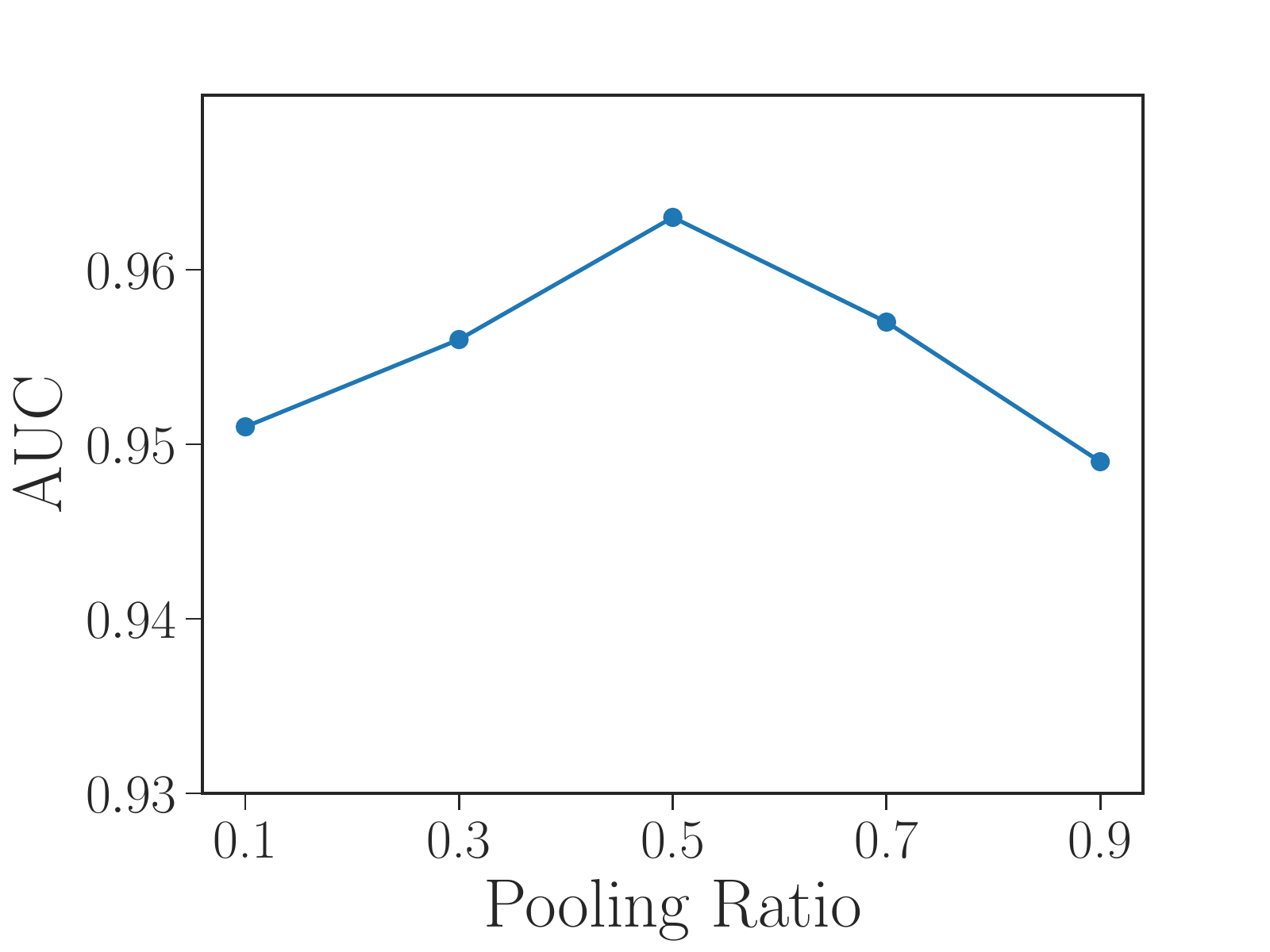}
            \caption[]%
            {}    
\vspace{-2mm}
            \label{fig:pool}
        \end{subfigure}
        \caption[ The average and standard deviation of critical parameters ]
        {Parameter sensitivity experiment results} 
        \label{fig:mean and std of nets}
\label{fig:param}
    \end{figure}

\vspace{1mm}
\noindent {\bf Results.}
As shown in Figure~\ref{fig:param}, overall, the impact of hyperparameter variation is insignificant.
Figure~\ref{fig:d_emb} shows that GoGNN reaches the best performance with representation dimension 128.
Figure~\ref{fig:d_hid} indicates that the salient point for the hidden feature size is 384.
As for the learning rate and pooling ratio, the best point appears at $10 \times 10^{-3}$ and $0.5$, respectively.

\section{Conclusion}
In this paper, we focus on structured entity interaction prediction.
This prediction demands the model to capture the information of the structure of entities and the interactions between entities.
However, the previous works represent the entities with insufficient information.
To address this limitation, we propose a novel model GoGNN which leverages the dual-attention mechanism in the view of graph of graphs to capture the information from both entity graphs and entity interaction graph hierarchically.
The experiments on real-life datasets demonstrate that our model could improve the performance on the chemical-chemical interaction prediction and drug-drug interaction prediction tasks.
GoGNN can be naturally extended to the applications on other graph of graphs such as financial networks, electrical networks, etc.
We leave the extension for future work.

\newpage
\bibliographystyle{named}
\bibliography{ref}

\end{document}